\documentclass[conference]{IEEEtran}
\IEEEoverridecommandlockouts
\usepackage{cite}
\usepackage{amsmath,amssymb,amsfonts}
\usepackage{algorithmic}
\usepackage{graphicx}
\usepackage{textcomp}
\usepackage{xcolor}
\def\BibTeX{{\rm B\kern-.05em{\sc i\kern-.025em b}\kern-.08em
    T\kern-.1667em\lower.7ex\hbox{E}\kern-.125emX}}
\usepackage{subfigure}
\usepackage{threeparttable}
\begin{document}

\title{Analysis of Sampling Strategies for Implicit 3D Reconstruction}

\author{\IEEEauthorblockN{1\textsuperscript{st} Qiang Liu}
\IEEEauthorblockA{\textit{School of Artificial Intelligence} \\
\textit{Jilin University}\\
Changchun, China \\
liuqiang5799@163.com}

\and

\IEEEauthorblockN{2\textsuperscript{nd} Xi Yang}
\IEEEauthorblockA{\textit{School of Artificial Intelligence} \\
\textit{Jilin University}\\
Changchun, China \\
yangxi21@jlu.edu.cn}

}

\maketitle

\begin{abstract}

In the training process of the implicit 3D reconstruction network, the choice of spatial query points' sampling strategy affects the final performance of the model. Different works have differences in the selection of sampling strategies, not only in the spatial distribution of query points but also in the order of magnitude difference in the density of query points. For how to select the sampling strategy of query points, current works are more akin to an enumerating operation to find the optimal solution, which seriously affects work efficiency. In this work, we explored the relationship between sampling strategy and network final performance through classification analysis and experimental comparison from three aspects: the relationship between network type and sampling strategy, the relationship between implicit function and sampling strategy, and the impact of sampling density on model performance. In addition, we also proposed two methods, linear sampling and distance mask, to improve the sampling strategy of query points, making it more general and robust.

\end{abstract}


\section{Introduction}
Fast and efficient 3D reconstruction technology has broad application prospects in virtual reality, autonomous driving, and other fields. In recent years, learning-based 3D reconstruction methods have become more and more popular because rich prior information of the 3D shape space can be encoded through deep learning. In addition to the explicit 3D reconstruction, researchers also explored implicit neural representations, which can generate the shape of 3D objects at any resolution, greatly reducing memory consumption in the training process.

\begin{figure*}[htbp]
    \centering  
	\subfigure[input object]{
        \label{fig:11}
		\includegraphics[width=0.21\linewidth]{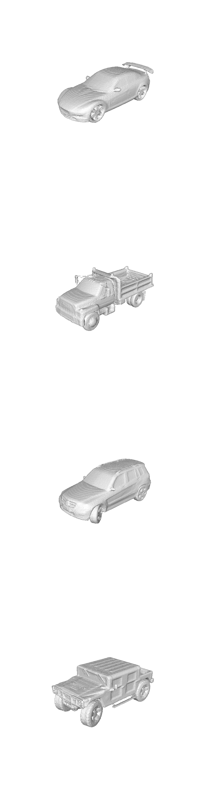}}
	\subfigure[uniform sampling]{
        \label{fig:12}
		\includegraphics[width=0.21\linewidth]{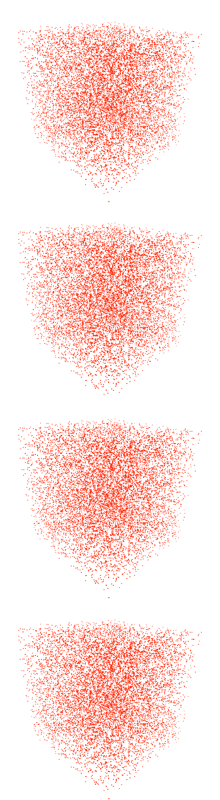}}
	\subfigure[surface sampling]{
        \label{fig:13}
		\includegraphics[width=0.21\linewidth]{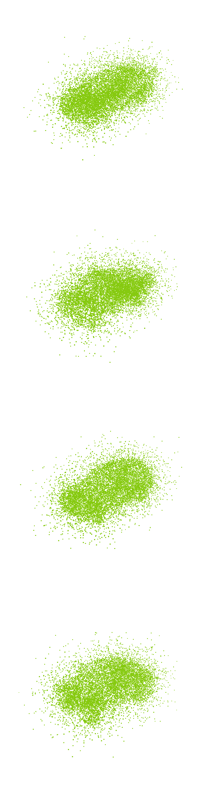}}
    \subfigure[hybrid sampling]{
        \label{fig:14}
		\includegraphics[width=0.21\linewidth]{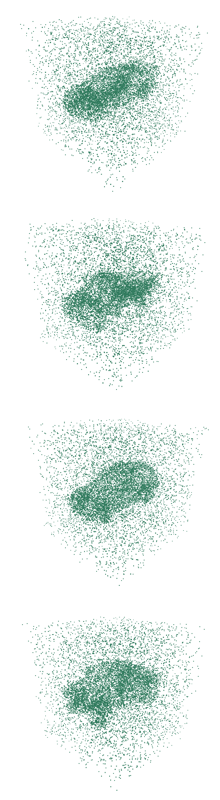}}	
    \caption{In the training process of implicit 3D reconstruction network, different strategies can be selected on how to sample the spatial query points of each object. (b) sample points randomly in the boundary space, (c) sample points near the surface of the object, and (d) mixes (b) and (c).}
    \label{fig:1}
\end{figure*}

The task of implicit 3D reconstruction is to learn a function to predict the spatial property of a query point at any position of the boundary space. We observe that in the network's training process, various implicit networks adopt different strategies to sample query points, which will be used for the learning of the model, such as uniform sampling \cite{onet, convonet}, surface sampling \cite{if-net, ndf, air-nets}, or hybrid sampling \cite{deepsdf, c-deepsdf, pifu, work6}, as shown in Fig. \ref{fig:1}. Different sampling strategies will bring different effects to the model. Some works have found this difference \cite{onet, pifu, air-nets}, but they did not explore the causes in depth.

In previous works, how to select the sampling strategy is more similar to an enumerating operation, which seriously reduces the work efficiency. Therefore, we try to explore the impact of different sampling strategies on implicit 3D reconstruction in this work. We hope to find the hidden laws based on the evaluation results through differentiated comparative experiments and explore the causes of these laws, so as to provide a guiding explanation for future related works to improve their work efficiency.

We classify current mainstream implicit reconstruction neural networks by studying the similarities and differences in network structures. In our view, the relevant networks can be divided into three types, namely MLP-Type, Convolution-Type, and Auto-Decoder-Type. By studying the performances of sampling strategies on different network architectures, we find that different types of models have different preferences on how to sample query points. The MLP-Type network is more robust to the change of sampling method. For Convolution-Type networks, it is not recommended to use near surface sampling strategy to avoid a significant decline in network performance, because it's necessary to obtain features of boundary-biased coordinates when embedding query points into feature space. By reason of the network characteristics of Auto-Decoder-Type networks, the strategy of uniform-biased sampling, which represents complete or close to uniform sampling, will make the network ineffective. Balanced surface sampling is usually a good choice and has a good evaluation result on various types of networks.

Different implicit functions will also affect the performance of the sampling strategy and we discuss this phenomenon. In general, for neural networks, the classification task is simpler than the regression task, and so is the extension to implicit 3D reconstruction. When the sampling strategy changes, the Occupancy implicit function will make the performance of the neural network more stable than the Signed Distance Function. Therefore, we recommend turning the implicit 3D reconstruction neural network into a classification task, which can not only ensure the final performance of the model but also improve the fault tolerance of the sampling strategy for query points.

In common sense, it seems that the more samples of query points for training, the better the neural network will learn. However, our experiments show that the number of sampling points in the boundary space does not have a decisive impact on the final effect of the model. Dense sampling increases the cost of training but does not necessarily improve the network performance. Therefore, it is important to control the number of sampling points, so that they can give consideration to both the training cost and the final effect of the work.

Based on the connection between the sampling strategy of the query point and the implicit 3D reconstruction, we propose two methods to improve the sampling strategy.  Firstly, we propose a sampling strategy based on linear distribution, which assigns sampling weights to spatial query points according to their distance from the object surface.  Linear sampling strategy ensures the stability of model performance and prevents the degradation of network performance due to inappropriate sampling methods, so it is a more general and robust sampling strategy.  Secondly, we propose a sampling strategy based on distance mask.  According to the input shape, a mask matrix is generated in the space, and the query point determines whether to enter the network calculation by obtaining the corresponding mask.  Although this method cannot help the network to better learn the fine reconstruction of shape, it can help to forcibly correct the parts prone to reconstruction errors, which ensures the reconstruction quality of objects to a large extent.

In summary, our contributions are as follows:

\begin{itemize}
\item We classify the implicit 3D reconstruction neural networks into three types and explore the impact of network types on the sampling strategy.

\item We explore the impact of the difference of implicit functions on the sampling strategy and confirm that the classification function has more advantages.

\item We investigate the impact of the sampling density of query points on model performance.

\item We propose two improved sampling strategies, linear sampling and distance mask, which guarantee the compatibility of the model and sampling strategies.

\end{itemize}

\section{Related works}

\subsection{Implicit Representation of 3D Objects}

Explicit 3D shape representation methods (voxels \cite{voxel}, meshes \cite{mesh}, pointclouds  \cite{pointnet, pointnet++}) have insurmountable limitations, such as voxels consume a lot of memory, meshes are difficult to represent complex topological structures, and pointclouds are discrete. In order to avoid these limitations mentioned above, continuous implicit functions have become a rapidly developing research field of 3D objects representation. Unlike the previously mentioned explicit representation, neural implicit representations can fit a continuous surface, generate it at any resolution, handle complex shape topologies naturally, and reduce the cost of the training process. When generating 3D objects, the mesh can be extracted from the learned model using algorithms such as Marching Cubes \cite{mcubes}, Multiresolution IsoSurface Extraction \cite{onet}, etc.

The first outstanding achievements in implicit 3D reconstruction are \cite{im-net, pifu, onet, deepsdf, disn}, and a large number of follow-up works \cite{if-net, convonet, air-nets, c-deepsdf, work1, work2, work3, work4, work5, work6, work7, work8, work9, work10, rgbd} continued to promote the development of implicit 3D reconstruction.

Mescheder \textit{et al.} \cite{onet} encoded the input shape by the encoder and predicted the occupancy function by the decoder with residual blocks \cite{resnet}. Park \textit{et al.} \cite{deepsdf} predicted the signed distance function based on a probabilistic auto-decoder. Peng \textit{et al.} \cite{convonet} and Chibane \textit{et al.} \cite{if-net} combined the complementary strengths of convolutional neural networks with those of implicit representations. Giebenhain \textit{et al.} \cite{air-nets} proposed a novel attention-based set abstraction method and an attentive decoder in conjunction with latent shape representation. Nam \textit{et al.} \cite{3dldm} proposed a diffusion model for neural implicit representations of objects that operates in the latent space of an auto-decoder.

In addition to continuing to optimize the neural network, researchers also explored new neural implicit representations. Chibane \textit{et al.} \cite{ndf} expressed the shape of an object by predicting the unsigned neural distance field, which reduces the requirements for watertight objects. Ye \textit{et al.} \cite{gifs} no longer divided 3D space into predefined categories, but focused on the relationship between every two 3D points, defining a binary flag to indicate whether two points are on the same side of the object surface, thus solving the problem of implicit representation of non-watertight or multiple objects. Ren \textit{et al.} \cite{geoudf} proposed a geometry-guided method for UDF and its gradient estimation that explicitly formulates the unsigned distance of a query point as the learnable affine averaging of its distances to the tangent planes of neighboring points.

There are also works focused on better converting 3D objects from implicit to explicit representations. Vetsch \textit{et al.} \cite{neuralmeshing} proposed a novel differentiable meshing algorithm for extracting surface meshes from neural implicit representations.

\subsection{Impact of Sampling Strategies}

Some implicit 3D reconstruction works have noticed that the sampling strategy will affect the final performance of the neural network. Differences in these strategies are briefly discussed \cite{onet, pifu, air-nets}, which involve uniform sampling \cite{onet, convonet} (selecting a certain number of points in the boundary space randomly), equal sampling \cite{onet} (selecting the same number of points inside and outside the 3D object),  surface sampling \cite{if-net, ndf, air-nets} (selecting points on the isosurface of the 3D object and applying one or several Gaussian noises with a standard deviation) and hybrid sampling \cite{deepsdf, c-deepsdf, pifu} (combining some of the above sampling methods in a certain proportion).

Mescheder \textit{et al.} \cite{onet} compared uniform sampling, equal sampling, and hybrid sampling, then found that the simplest uniform sampling had the best effect. Then they explained that other sampling strategies would cause deviation to the model, for example, equal sampling would implicitly tell the model that the volume of each object was 0.5. They also compared the impact of the density of query points. When the number of query points for each object is reduced from 2,048 to 64 in the training, the model performance is only slightly degraded. 

Shunsuke \textit{et al.} \cite{pifu} also compared different sampling methods and explained that if the points in 3D space are uniformly sampled, most of the points are far away from the isosurface, which will unnecessarily weight the network to external prediction. If only sampling around the isosurface, it may lead to overfitting, Therefore, it is recommended to combine uniform sampling with adaptive sampling based on surface geometry. 

Giebenhain \textit{et al.} \cite{air-nets} found that the two datasets of surface sampling and uniform sampling were used respectively, and the latter made the neural network perform better in general. Therefore, it was speculated that the uniform distribution query might be more applicable to all models.

\section{Implicit 3D Reconstruction}

\subsection{Key Idea}

Although works related to implicit 3D reconstruction in recent years are different in the selection of network structure and implicit function, they are similar in concept. The key idea of implicit 3D reconstruction is that the network is given a latent vector $\textbf{\emph{z}}$ representing shape features and the query point $\textbf{\emph{p}}$ in the boundary space, and then outputs the spatial property of the query point,

\begin{equation}
f_{\theta}(\textbf{\emph{z}}, \textbf{\emph{p}}) \longmapsto pro(\textbf{\emph{p}})
\end{equation}

where $\textbf{\emph{z}}$ can be encoded by the encoder or randomly assigned to the neural network at the beginning of training, and $pro(\textbf{\emph{p}})$ represents the spatial property of the query point $\textbf{\emph{p}}$ relative to the 3D object, such as inside/outside the object, the distance to the object surface, etc.

\subsection{Data Preprocessing}

Data preprocessing is divided into two parts: generating 3D shape representation and generating spatial query points.

The former is to directly collect points on the surface of 3D objects, and works \cite{if-net, ndf} may require a further operation to generate discrete voxel grids based on KD-Tree.

The latter is our focus. The three coordinate values of each query point in the 3D space are all get from random values in the range of $[-0.5,0.5]$, so as to achieve uniform sampling. Surface sampling can be generated by adding Gaussian noise with a certain standard deviation $\sigma$ to the points collected from the isosurface of the 3D object. Hybrid sampling is a proportional concatenation of several query point sets.

\subsection{Sampling Strategies \label{sec.smp}}

In the training process of the network, we can observe the differences in the strategies of sampling query points in different work from the horizontal and vertical aspects.

\begin{figure}[t]
\centerline{\includegraphics[width=\linewidth]{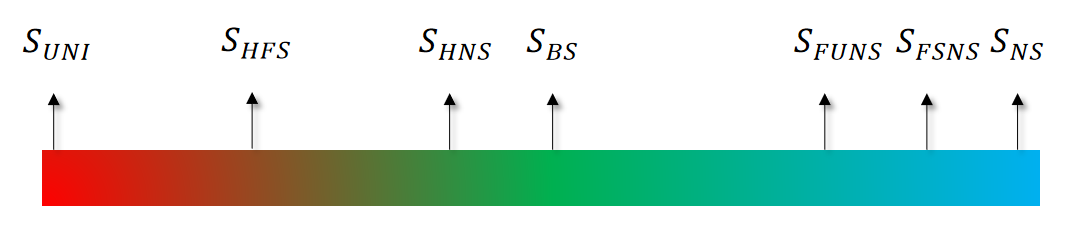}}
\caption{We select 7 sampling strategies, from uniform sampling to surface sampling, to explore the influence of the spatial distribution of query points on the network performance.}
\label{fig:2}
\end{figure}

Horizontally, the spatial distribution of query points is different. We think it's similar to a gradual spectrum. One end is fully uniform sampling in the boundary space, and the other end is pure surface sampling on the object's isosurface. \cite{importance} provides a sampling function to adjust the spatial distribution of query points by changing the threshold of the function, but most works control the distribution of sampling points by combining different sampling methods. In this work, we select a total of 7 sampling strategies for comparative experiments, as shown in Fig. \ref{fig:2}.

\begin{figure*}[t]
    \centering
    \subfigure[MLP-Type]{
    \label{fig:cls.1}
    \includegraphics[width=0.352\linewidth]{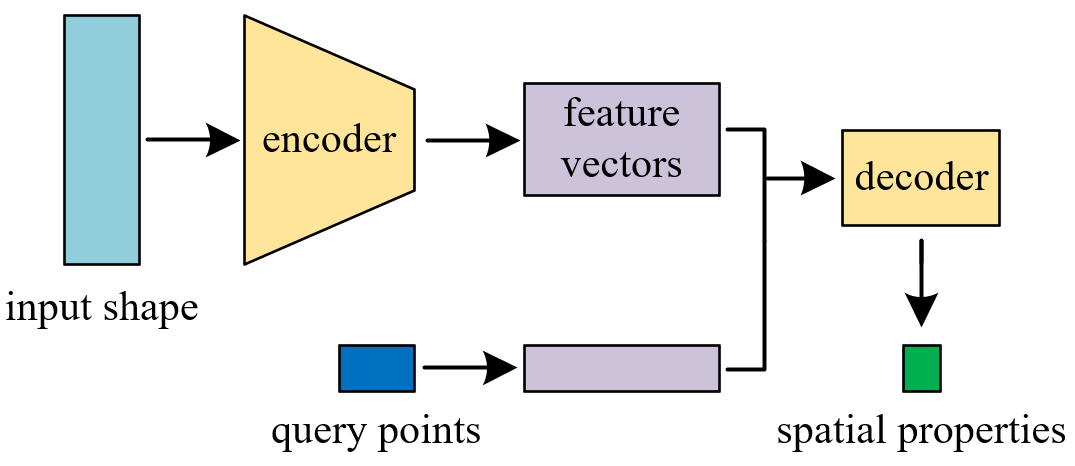}}
    \subfigure[Convolution-Type]{
    \label{fig:cls.2}
    \includegraphics[width=0.352\linewidth]{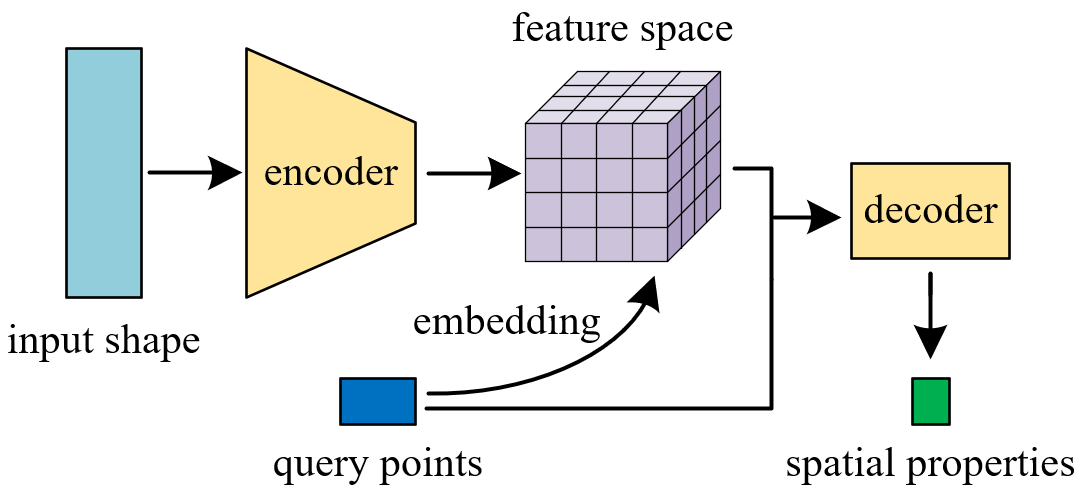}}
    \subfigure[Auto-Decoder-Type]{
    \label{fig:cls.3}
    \includegraphics[width=0.261\linewidth]{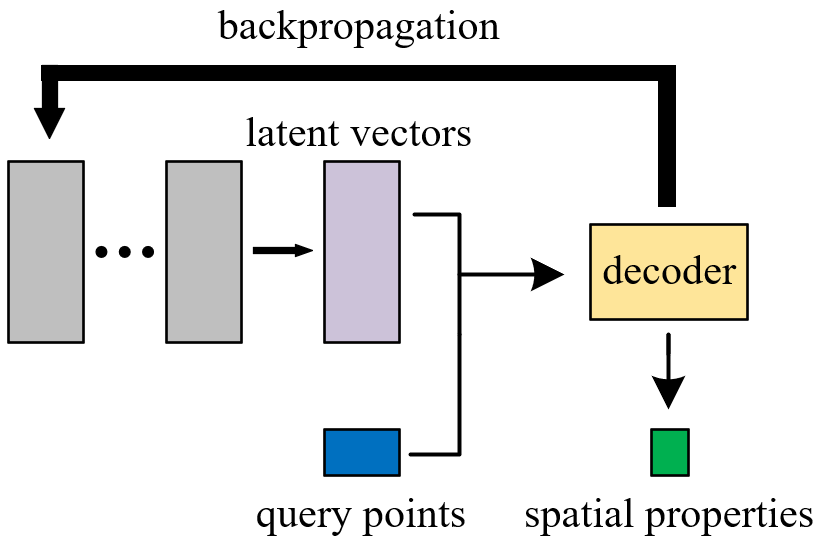}}
    \caption{Based on how to obtain features of the input shape and how to combine the coordinates of query points with the above shape features in network structure, we classify these implicit 3D reconstruction models into three categories.}
\end{figure*}

\begin{itemize}
    \item $S_{UNI}$: Uniform Sampling.
    
    \item $S_{HFS}$: Hybrid Far Surface Sampling. Half of the points are obtained from $S_{UNI}$, and the remaining half are obtained from surface sampling plus some Gaussian noise with $\sigma=0.1$.
    
    \item $S_{HNS}$: Hybrid Near Surface Sampling. Similar to $S_{HFS}$, except that the standard deviation of Gaussian noise is $0.01$.
    
    \item $S_{BS}$: Balanced Surface Sampling. Surface sampling with $\sigma=0.1$ and $\sigma=0.01$ are taken 50$\%$ each.
    
    \item $S_{FUNS}$: Few Uniform - Near Surface Sampling. Adding 1$\%$ $S_{UNI}$ to $S_{NS}$.
    
    \item $S_{FSNS}$: Few Surface - Near Surface Sampling. Adding 1$\%$ surface sampling with $\sigma=0.1$ to $S_{NS}$.
    
    \item $S_{NS}$: Near Surface Sampling. Surface sampling with $\sigma=0.01$ and $\sigma=0.001$ are taken 50$\%$ each.
    
\end{itemize}

Vertically, different sampling densities are adopted. \cite{onet, convonet, air-nets} adopt sparse sampling, and \cite{if-net, ndf, deepsdf, c-deepsdf} choose dense sampling. In this work, we select sparse sampling (2,000 points) and dense sampling (20,000 points) for each object. The following experimental results, unless otherwise specified, default to sparse sampling.

\subsection{Experimental Details \label{sec.exp}}
\textbf{Dataset}: In this work, we follow \cite{mesh_fusion} to make the ShapeNet watertight. In order to speed up the experiment, we only select about 3,000 objects of cars. Therefore, the experimental results may not be consistent with the data provided in the original work, but our purpose is to explore the impact of the sampling strategy on the experimental results, not to pursue the optimal performance of the model.

\textbf{Networks}: According to different network designs, we select 6 models for experiments, including DeepSDF \cite{deepsdf} and Curriculum-DeepSDF \cite{c-deepsdf} with $S_{FUNS}$, ONet \cite{onet} and ConvONet \cite{convonet} with $S_{UNI}$, and IF-Net \cite{if-net} and AIR-Nets \cite{air-nets} with $S_{BS}$. The above sampling strategies are used in the original works. In order to facilitate comparison, we only adopt the inputs of pointclouds for all networks and do not explore images inputs or voxels inputs. In addition, some works add Gaussian noise with a small standard deviation to the pointclouds inputs, while others do not. Here, Gaussian noise is not adopted for all inputs. Except for Sec \ref{sec.func}, the implicit functions we choose all follow the original work of each model.

\section{Analysis of Sampling Strategy}

In Sec \ref{sec.cls}, we classify current implicit 3D reconstruction neural networks into three categories according to the similarities and differences in network structures, and in Sec \ref{sec.net}, we verify the rationality of our network classification and discuss the performance differences when selecting different sampling strategies of query points under different network types.

In Sec \ref{sec.func}, we discuss the impact of the change of implicit functions on the evaluation results of different sampling strategies and give a solid suggestion that classification implicit functions are more worthy of selection.

In Sec \ref{sec.dens}, we also verify that dense sampling will not greatly improve the network performance but increase the training cost. Sparse sampling with thousands of samples is enough to ensure the performance of the network model.

\subsection{Network Type \label{sec.cls}}

\textbf{We observe that the focus of implicit 3D reconstruction neural networks is how to obtain features that characterize the shape of the input, and how to combine the coordinates of query points with shape features for decoding operations. Based on the above two aspects, we classify implicit neural networks into three types}.

\textbf{MLP-Type}: Although there are some differences in MLP-Type networks' \cite{im-net, onet, air-nets} design, they are similar in our view. MLP-Type networks adopt a typical Encoder-Decoder-Type structure. At the encoder and decoder stages, the original network design may not be limited to the connection of multiple linear layers, but there may also be some convolution layers with kernel-size of 3 and stride of 1. The latter is equivalent to the former, so we call all of them MLP-Type networks. In a word, MLP-Type networks follow such an idea that the global or local features of the input shape are obtained through MLP and combined with the feature vectors of the spatial query points. Finally, the spatial property of query points are output through MLP. As shown in Fig. \ref{fig:cls.1}.

\textbf{Convolution-Type}: Convolutional Neural Networks (CNNs) have achieved extraordinary achievements in many two-dimensional tasks, including but not limited to image classification and segmentation. Some works \cite{convonet, if-net, ndf, gifs, work5, work6, work9, work10} have tried to introduce CNNs into implicit 3D reconstruction. For this reason, although Convolution-Type networks also use the Encoder-Decoder structure, it is different from MLP-Type networks. In order to use the multi-layer receptive field of CNNs, it is necessary to obtain the two-dimensional or three-dimensional feature space of the 3D object, and then use bilinear/trilinear interpolation to embed the query points into the above feature space according to the coordinate position of these points, so as to obtain the feature vector representing the relationship between the query points and the shape space, as shown in Fig. \ref{fig:cls.2}. The final decoding stage is similar to MLP-Type networks, where multiple linear layers map the previously obtained feature vectors to the spatial property of the input query points.

\textbf{Auto-Decoder-Type}: As shown in Fig. \ref{fig:cls.3}, different from the above two types that need to obtain shape features through the encoder, \cite{deepsdf, c-deepsdf, work7} adopt the encoder-less setting. An initial latent vector is generated for each object in the experiment. In the training process, not only the decoder weight is modified through backpropagation, but also the latent vector representing the object shape is continuously optimized. In the generation process, the weights of the decoder are fixed, then sampling points and their corresponding ground truth are passed in. Subsequently, corresponding shape latent vectors are modified through continuous iteration. After the iteration, the spatial property of the dense sampling points in the entire boundary space is predicted.

\subsection{Impact of Network Type on Sampling Strategies \label{sec.net}}

\textbf{Network type has an important impact on the selection of sampling strategy. When facing changes in sampling strategy, MLP-Type networks are robust, Convolution-Type networks are not friendly to near surface sampling, and Auto-Decoder-Type networks are not suitable for uniform-biased sampling.} 

\begin{figure}[t]
	\centering
	\subfigure[F-Score($1.5\%$)]{
		\label{fig:eval.1}
		\includegraphics[width=1.0\linewidth]{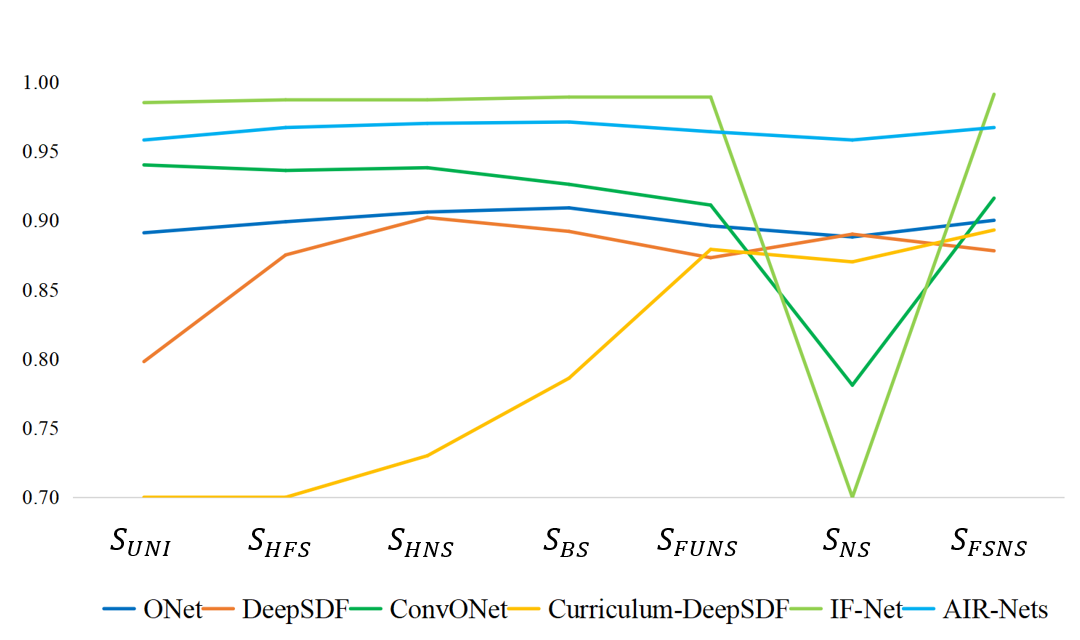}}
	\subfigure[Normal Consistency]{
		\label{fig:eval.2}
		\includegraphics[width=1.0\linewidth]{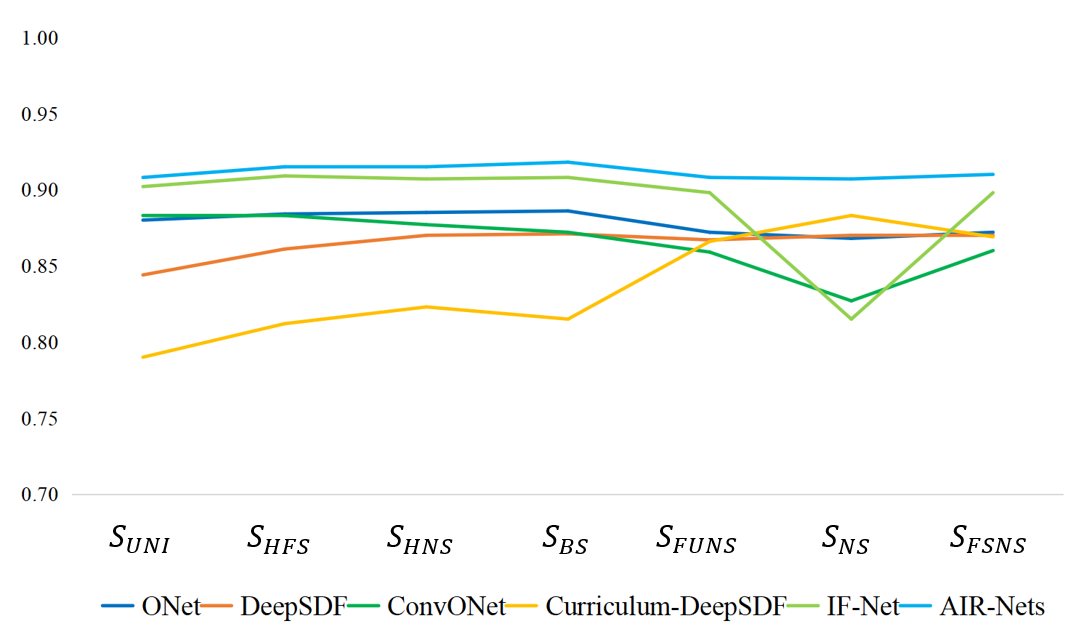}}
	\caption{Evaluation results of F-Score($1.5\%$) (a) and Normal Consistency (b) under various sampling strategies in different networks. Different colors represent the evaluation results of different types of networks. Blue represents MLP-Type networks, green represents Convolution-Type networks, and yellow represents Auto-Decoder-Type networks.}
        \label{fig:eval}
\end{figure}

According to the analysis in Section \ref{sec.cls}, we think that the 6 models we selected can be summarized respectively, that is, ONet \cite{onet} and AIR-Nets \cite{air-nets} belong to MLP-Type networks, IF-Net \cite{if-net} and ConvONet \cite{convonet} belong to Convolution-Type networks, and DeepSDF \cite{deepsdf} and Curriculum-DeepSDF \cite{c-deepsdf} belong to Auto-Decoder-Type networks. For each model, we use the 7 sampling strategies in Sec \ref{sec.smp} to observe whether the evaluation results of similar networks are correlated. 

The final evaluation results of F-Score($1.5\%$) and Normal Consistency are shown in Fig. \ref{fig:eval.1} and Fig. \ref{fig:eval.2}. Due to the poor results of some networks under certain sampling strategies, we truncate the results below the minimum value of the vertical axis of the coordinate to take into account the observational effects of other model evaluations.

Just as in our classification, there are significant differences in evaluation results in different types of networks, while there are similarities among the same type of networks. Different from other networks, yellow Auto-Decoder-Type networks perform poorly when uniform-biased sampling is adopted. Under near surface sampling, green Convolution-Type networks show obvious differences, and their performance plummets. Then we will analyze the reasons from these two aspects.

\subsubsection{Uniform-biased Sampling Is Not Applicable to Auto-Decoder-Type Networks}

Obviously, there is a visible difference between Encoder-Decoder-Type networks and Auto-Decoder-Type networks. For the former, uniform sampling usually ensures good results, although it may not be the best. For the latter, when uniform-biased sampling is adopted, the performance of the model is poor. It is worth noting that the implicit function chosen here for Auto-Decoder-Type networks is the Signed Distance Function (SDF), while other networks use the Occupancy (Occ). When the Truncated Signed Distance Function (TSDF) is adopted for Auto-Decoder-Type networks following the original work \cite {deepsdf, c-deepsdf}, the model will face the risk of failure, which is discussed in Sec \ref{sec.func}. It is obvious that the response of Auto-Decoder-Type networks to the change in sampling strategies is different from that of other types of neural networks. 

We propose the causes of this phenomenon. Since Auto-Decoder-Type networks adopt an encoder-less setting, they cannot extract features from the input shape by an encoder, so they can only randomly assign a latent vector to input data at the beginning of the experiment and continuously optimize it in the training process. Therefore, the spatial distribution of the query points input is crucial for learning the shape of the object. Taking a large proportion of sample points near the 3D object's surface can better approach the shape, and the learning effect will be better. Curriculum-DeepSDF \cite{c-deepsdf} has higher requirements for surface sampling and is more sensitive to fluctuations as a result of multi-level surface accuracy and sample difficulty being considered in its network design. 

Compared with other networks, Auto-Decoder-Type networks present a different trend under different sampling strategies: the performance of the network generally improves with the increase of surface sampling proportion, due to the feature vector used to represent the object shape depending on how the spatial query points are sampled. Naturally, surface sampling should dominate the whole sampling process.

\begin{figure}[t ]
	\centering
	\subfigure[ONet]{
		\label{fig:recons.1}
		\includegraphics[width=0.48\linewidth]{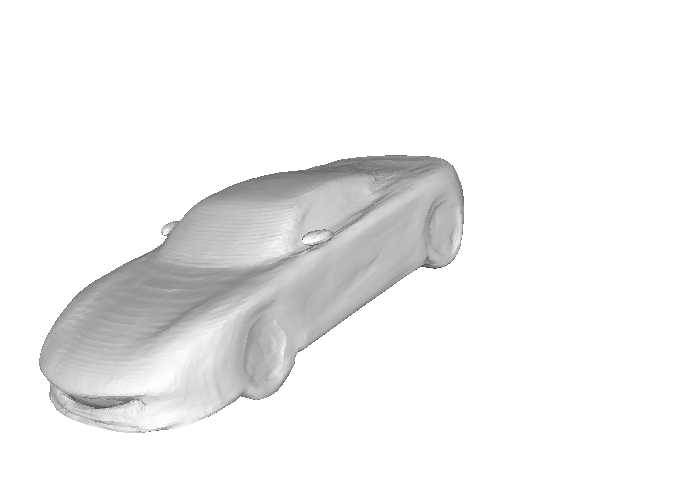}}
	\subfigure[AIR-Nets]{
		\label{fig:recons.2}
		\includegraphics[width=0.48\linewidth]{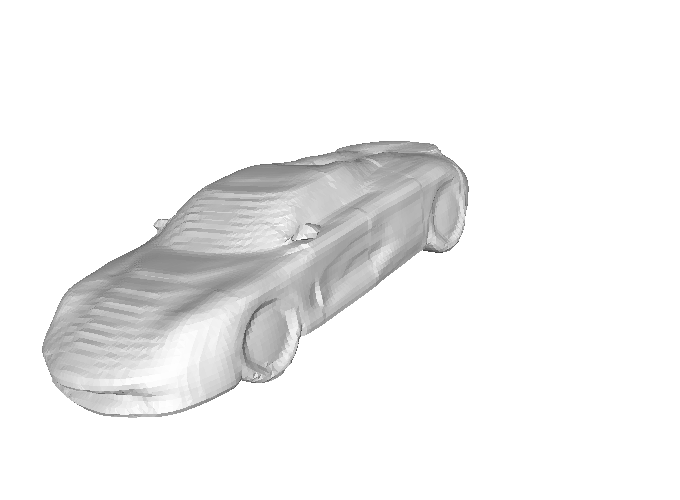}}
	\subfigure[IF-Net]{
		\label{fig:recons.3}
		\includegraphics[width=0.48\linewidth]{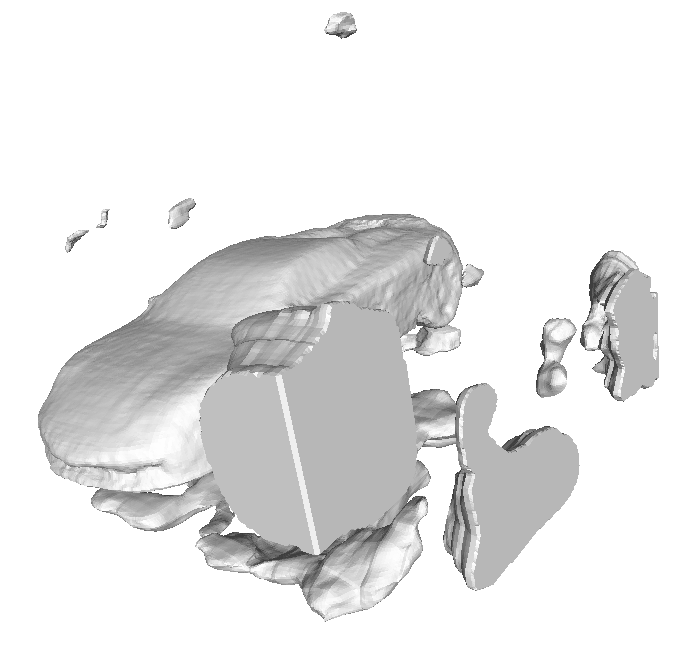}}
        \subfigure[ConvONet]{
		\label{fig:recons.4}
		\includegraphics[width=0.48\linewidth]{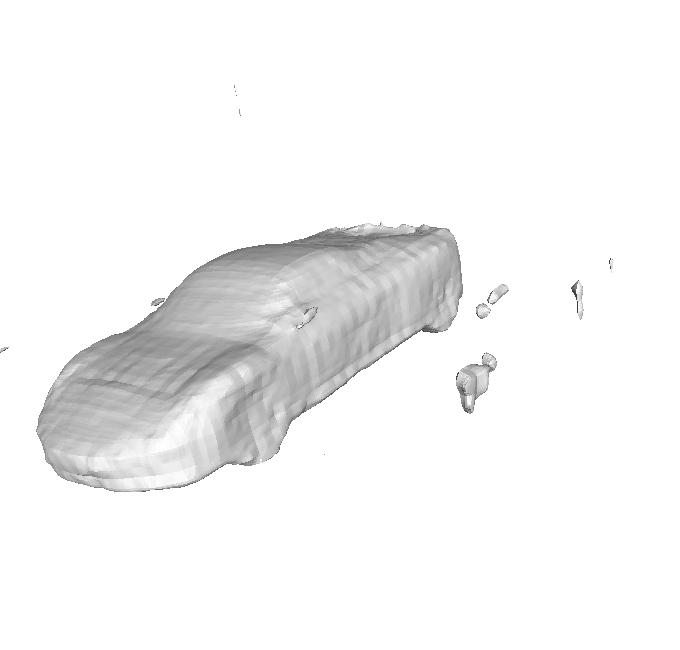}}
	\caption{When $S_{NS}$ is adopted, different type networks show diverse performance: MLP-Type (a) and (b) can still reconstruct the object shape well, and the spatial property of query points can be basically predicted correctly in the whole boundary space. Convolution-Type (c) and (d) can be correctly predicted near the surface of the object, while the position far away from the object is wrong.}
        \label{fig:recons}
\end{figure}

\begin{table*}[t]

\caption{Evaluation results of F-Score(1.5$\%$) on sampling strategies with different implicit functions. Evaluation results under each implicit function have calculated the changes compared with Occ.}
\begin{center}
 \resizebox{\linewidth}{!}{   
\begin{threeparttable}

    \begin{tabular}{|c|c|c|c|c|c|c|c|}
    \hline
    networks $\&$ implicit functions & $S_{UNI}$ & $S_{HFS}$ & $S_{HNS}$ & $S_{BS}$ & $S_{FUNS}$ & $S_{NS}$ & $S_{FSNS}$ \\
    \hline
    ONet $\&$ Occ & \textbf{0.89} & \textbf{0.90} & \textbf{0.91} & 0.91 & \textbf{0.90} & \textbf{0.89} & \textbf{0.90} \\
    ONet $\&$ SDF & 0.86 ($\downarrow 3.4\%$) & 0.89 ($\downarrow 1.1\%$) & 0.89 ($\downarrow 2.2\%$) & \textbf{0.92} ($\uparrow 1.1\%$) & 0.67 ($\downarrow 25.6\%$) & 0.74 ($\downarrow 16.7\%$) & 0.62 ($\downarrow 31.2\%$)  \\
    ONet $\&$ TSDF & 0.87 ($\downarrow 2.2\%$) & 0.89 ($\downarrow 1.1\%$) & 0.90 ($\downarrow 1.1\%$) & 0.90 ($\downarrow 1.1\%$) & 0.85 ($\downarrow 5.6\%$) & 0.84 ($\downarrow 5.6\%$) & 0.67 ($\downarrow 25.6\%$) \\
    
    \hline
    IF-Net $\&$ Occ & \textbf{0.99} & \textbf{0.99} & \textbf{0.99} & \textbf{0.99} & \textbf{0.99} & \textbf{0.32} & \textbf{0.99} \\
    IF-Net $\&$ SDF & \textbf{0.99} & 0.98 ($\downarrow 1.0\%$) & \textbf{0.99} & \textbf{0.99} & 0.98 ($\downarrow 1.0\%$) & 0.16 ($\downarrow 50.0\%$) & \textbf{0.99} \\
    IF-Net $\&$ TSDF & \textbf{0.99} & \textbf{0.99} & \textbf{0.99} & \textbf{0.99} & \textbf{0.99} & 0.24 ($\downarrow 25.0\%$) & \textbf{0.99} \\
    
    \hline
    DeepSDF $\&$ Occ & / & / & / & 0.87 & \textbf{0.90} & \textbf{0.89} & \textbf{0.88} \\
    DeepSDF $\&$ SDF & \textbf{0.80} & \textbf{0.88} & \textbf{0.90} & 0.89 ($\uparrow 2.3\%$) & 0.87 ($\downarrow 3.3\%$) & \textbf{0.89} & \textbf{0.88} \\
    DeepSDF $\&$ TSDF & / & / & \textbf{0.90} & \textbf{0.90} ($\uparrow 3.4\%$) & 0.88 ($\downarrow 2.2\%$) & 0.87 ($\downarrow 2.2\%$) & \textbf{0.88} \\
    
    \hline
    \end{tabular}
    
\end{threeparttable}}
\end{center}
\label{tab:func}
\end{table*}

\subsubsection{Near Surface Sampling Is Not Suitable for Convolution-Type Networks }

We can see in Fig. \ref{fig:eval} that Convolution-Type networks are more sensitive and not friendly to $S_{NS}$ compared with robust MLP-Type networks. However, when only $1\%$ of the query points which are get from uniform-biased sampling are added, such as $S_{FUNS}$ and $S_{FSNS}$, the network performance can be recovered.

With Convolution-Type networks having an additional process of interpolating the shape feature space according to the spatial coordinates of the query points, the feature vectors representing the query points are closely connected with their coordinates. When $S_{NS}$ is adopted, the feature space vectors corresponding to the boundary-biased points are not collected in the whole training process, Therefore, when the sampling points in the whole space are input during generation, the network cannot correctly predict the boundary-biased points of the space, and it is difficult to reconstruct the 3D shape with high quality. From the reconstruction effect, the positions near the object can basically be recovered correctly, while the parts far away from the object are predicted incorrectly, as shown in Fig. \ref{fig:recons}. That is because $S_{NS}$ only collects features nearby the object in the feature space, it is difficult to take into account the boundary-biased position. 

This also explains that $S_{FUNS}$ and $S_{FSNS}$ are only a few more boundary-biased points than $S_{NS}$, but the network performance can be greatly improved. The corresponding features of the boundary-biased points in the feature space are largely the same, and then the features obtained by interpolation are the same, so only a small number of the boundary-biased points are needed to obtain better results.

%% \subsubsection{How to Select Sampling Strategies for Different Types of Networks}

In summary, different network designs have an important impact on the selection of sampling strategies. MLP-Type networks' performance is robust, and different sampling strategies have little difference in evaluation results. In order to ensure the performance of the model without large deviation, Convolution-Type networks should avoid near surface sampling, and Auto-Decoder-Type networks should avoid uniform-biased sampling. Both query points of surface-biased and boundary-biased should all be taken into account, and balanced surface sampling is most effective in almost all networks.

\subsection{Impact of Implicit Function on Sampling Strategies \label{sec.func}}

\textbf{The classification implicit function has the higher fault tolerance for different sampling strategies and will be a better choice.}

Another key point of implicit 3D reconstruction is to select an appropriate implicit function, that is, to select the spatial relationship between query points and the 3D shape. In our view, in implicit 3D reconstruction, the selection of implicit function is largely related to the selection of neural networks task. Therefore, just like to classify the tasks of neural networks, we divide various implicit functions into two categories: classification implicit functions (Occ, GIFS, etc.) \cite{im-net, onet, if-net, convonet, air-nets, gifs, work5} and regression implicit functions (SDF, TSDF, UDF, CSP, etc.) \cite{deepsdf, c-deepsdf, ndf, work3, work6, work7, work8}.

We find in Tab \ref{tab:func} that some sampling methods of DeepSDF \cite{deepsdf} under Occ and TSDF will make the network invalid, which is also a common problem for Auto-Decoder-Type networks. When the sampling strategy is uniform-biased sampling, as a result of the signed distance being truncated, a large part of the TSDF value becomes a constant truncation distance $\tau$, since the points far away from the object always occupy the majority of space. This is a very bad result for Auto-Decoder-Type networks because the final feature vector representing the shape almost entirely depends on how to provide the network with spatial query points in the continuous and iterative optimization process. If the uniform-biased sampling is adopted, the function represented by the neural network will degenerate into a constant function, because a large number of query points' corresponding ground truths are marked as $\tau$. We output the SDF values predicted by the network in this case, which is basically floating up and down $\tau$, consistent with our conjecture. This reason can also be extended to the Auto-Decoder-Type network with the Occ selected. When most of the occupancy of query points is 0, the network tends to become a constant function with a value of 0.

TSDF certainly makes the network more focused on the vicinity of the object surface and is more stable than SDF. But we can clearly see in Table \ref{tab:func} that the most robust implicit function is Occ. In general, for neural networks, the classification problem can be seen as a special case of the regression problem. The output of 0.5 or 1 under the classification task seems to be very different, but it is consistent in meaning. But the difference of 0.1 in the regression task is already a big difference. We have reason to think that the selection of Occupancy as an implicit function is to take the neural network as a simple binary network, and the natural final performancet will be better.

To sum up, the change of implicit function does not change the evaluation trend among sampling strategies, but it may magnify this trend. In order to improve the fault tolerance of the implicit 3D reconstruction to the sampling strategy of the query points, we recommend the classification implicit function more than the regression implicit function which is more sensitive to the spatial distribution of the sampling points. This makes the network task simpler and has more stable results in the face of different sampling strategies.

\begin{table}[tb]   
    \caption{Effect of sampling density, where \textbf{\emph{E.S}} and \textbf{\emph{E.D}} represent \textbf{evaluation} results of IoU under \textbf{sparse sampling} and \textbf{dense sampling}, \textbf{\emph{M.M}} and \textbf{\emph{M.T}}  represent the \textbf{multiple} of \textbf{memory} and \textbf{time required for each iteration cycle} required for dense sampling under the same configuration compared to sparse sampling.}

    \begin{center}
    \begin{tabular}{|c|c|c|c|}
    \hline
     & ONet & IF-Net & DeepSDF\\
    \hline
    \textbf{\emph{E.S}} & \textbf{0.88} & \textbf{0.94} & 0.88 \\
    \textbf{\emph{E.D}} & 0.87 & 0.87 & \textbf{0.92} \\
    \hline
    \textbf{\emph{M.M}} & 2.85 & 1.49 & 6.27 \\
    \hline
    \textbf{\emph{M.T}} & 1.97 & 2.63 & 7.66 \\
    \hline
    \end{tabular} 
\end{center}  
\label{tab:dens}
\end{table}

\subsection{Impact of Sampling Density \label{sec.dens}}

\textbf{Dense sampling makes the training cost increase exponentially, but the model performance may not be improved accordingly. Sparse sampling is sufficient to meet the reconstruction requirements.}

We observe that some implicit 3D reconstruction works sparsely sample query points from space, while other works adopt dense sampling. Dense sampling will change the final performance of the model, but it is worth wondering whether it is all worth it given the additional experimental costs that come with it. In order to explore whether this difference will affect the training effect of the model, we evaluate the reconstruction effect of each network when sampling 2,000 and 20,000 points, respectively. Here, we explore the impact of sparse sampling and dense sampling on the evaluation results in 3 models: ONet \cite{onet}, IF-Net \cite{if-net}, and DeepSDF \cite{deepsdf}. Limited by space, only $S_{BS}$ which performs well in each network are shown here. The results are shown in Tab \ref{tab:dens}.

Obviously, the performance of the implicit 3D reconstruction neural network does not have a positive correlation with the sampling density but the increase of the sampling density will lead to a significant increase in the time and space cost in the experiment. Therefore, how to control the sampling number of spatial query points is an important problem, which relates to the cost of network training. In our opinion, thousands of samples are enough to make the network achieve optimal performance.

\section{Improvement of Sampling Strategy}

we realize that sampling strategy affects the model effect of implicit 3D reconstruction to a large extent based on the analysis in the previous chapter,  and the most important factor affecting the selection of sampling strategy is the type of network: Convolution-Type network and Auto-Decoder-Type network, because of their special network structure, the former is not suitable for near-surface sampling strategy, while the latter is not friendly to partial uniform sampling strategy.

Therefore, considering the incompatibility between implicit neural representation network and some sampling strategies leads to the degradation of model performance, we put forward two improvement ideas of "data pre-processing" and "data post-processing" to adjust the sampling strategies.

The idea of "data pre-processing" indicates that it is necessary to process the query points well before the neural network starts to learn, so as to prevent the wrong sampling method from leading to poor network learning effect. By setting an appropriate sampling strategy, we make the spatial distribution of the query points suitable for implicit neural representation network learning, so as to ensure that the network model can correctly learn the spatial attribute of any point in the boundary space.  The key to this idea is how to set up a new sampling strategy.

The idea of "data post-processing" means that regardless of whether the neural network can correctly predict the whole boundary space, only the wrong part will be processed after the model prediction. No matter whether the spatial distribution of the input query points is appropriate or not, the parts of the network that are easy to predict errors are screened out, and only the query points with correct prediction are retained, which ensures the reconstruction effect of the model to a large extent.   The key of this way of thinking is how to find out and shield the space query points which are easy to make mistakes.

Based on the above two ideas, in this chapter, we propose two improved sampling strategy algorithms based on linear distribution and distance mask, and carry out experimental evaluation on these methods.

\subsection{The Sampling Strategy based on Linear Distribution}

\subsubsection{Algorithm}

For the implicit 3D reconstruction network, the spatial query points as the learning sample in the training process needs to meet two conditions: one is to provide the necessary remote surface sampling points, so that the neural network can learn to determine the approximate decision boundary of the internal and external 3D objects in the boundary space, and realize the preliminary contour reconstruction. The proportion of query points in this part does not need to be large, but it is essential. The second is to provide a large proportion of near-surface sampling points, which is the key to implicit 3D reconstruction, so that the model can learn how to achieve fine surface reconstruction of 3D objects.

Therefore, we propose a linear sampling strategy ($S_{Linear}$). Considering that most of the points in the boundary space are at a medium distance from the surface of the 3D object, the points in the near distance are of great importance, while the points in the ultra-far distance do not need to be introduced into the neural network as query points. A distance threshold $\tau$ is set to control the sampling of these space points.

\begin{figure}[h]
\centerline{\includegraphics[width=\linewidth]{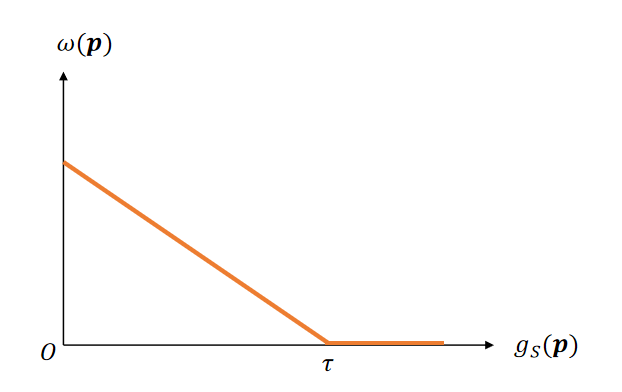}}
\caption{Within a distance threshold, the sampling weight of a point decreases linearly as the absolute value of the distance from the object surface increases.}
\label{fig:linear}
\end{figure}

If the absolute distance $g_S(\textbf{\emph{p}})$ between the collection point $\textbf{\emph{p}}$ and the shape surface $S$ does not exceed the threshold $\tau$, it is considered a necessary query point and is believed to be helpful for neural network learning how to reconstruct object shapes in boundary space. If $g_S(\textbf{\emph{p}})$ exceeds $\tau$, it will be considered as an unnecessary query point and will no longer be transmitted to the neural network. Therefore, as shown in Fig. \ref{fig:linear}, for points in space, we linearly decrease the assigned weight $\omega(\textbf{\emph{p}})$ as they increase. That is, the closer the spatial points are to the surface of the object, the higher the probability of being collected. On the contrary, the farther the spatial points are from the surface of the object, the lower the probability of being collected.

\begin{table*}[tb]   
    \caption{$D-Score$ for different sampling strategies.}
    \begin{center}
    \resizebox{0.7\linewidth}{!}{
    \begin{tabular}{|c|c|c|c|c|c|c|c|c|}
    \hline
     & $S_{UNI}$ & $S_{HFS}$ & $S_{HNS}$ & $S_{FUNS}$ & $S_{BS}$ & $S_{NS}$ & $S_{FSNS}$ & $S_{Linear}$\\
    \hline
    $D-Score$ & 0.140 & 0.106 & 0.065 & 0.079 & 0.059 & 0.226 & 0.057 & \textbf{0.049}\\
    \hline
    \end{tabular}}
    \end{center}
    
\label{tab:linear}
\end{table*}

In order to realize the linear sampling, we designed a linear function, according to the distance of the collection point relative to the shape surface, the corresponding collection weight of the point is determined.

\begin{equation}
    \omega(\textbf{\emph{p}}) = \left\{
    \begin{array}{ccl}
        \mu \cdot (\tau - g_S(\textbf{\emph{p}})) & , & {0 \leq g_S(\textbf{\emph{p}}) \leq \tau}\\
        0 & , & {g_S(\textbf{\emph{p}}) > \tau}
    \end{array} \right. 
\end{equation}

The distribution weight $\omega(\textbf{\emph{p}})$ of the spatial acquisition point $\textbf{\emph{p}}$ is jointly determined by the two parameters $\mu$, $\tau$ and the corresponding absolute value $g_S(\textbf{\emph{p}})$ of the distance from the relative shape surface $S$. $\tau$ determines how far away from the object's surface the assigned weight falls to 0. $\mu$ means that as the absolute value of the distance increases, the magnification of the weight linearly decreases, that is, the larger the value of $\mu$, the faster $\omega(\textbf{\emph{p}})$ decreases.

\subsubsection{Evaluation}

We use the linear sampling strategy $S_{Linear}$ to generate spatial query points with a number of 2,000 for each 3D object. We evaluate experimentally on the 6 implicit 3D reconstruction networks selected in Sec. \ref{sec.exp} and compare with the 7 sampling strategies selected in Sec. \ref{sec.smp}. It should be noted that, in our comparative experiments on each model, except for the different spatial distribution of the query points, the rest adopt the same experimental configuration in order to observe the evaluation effect of the linear sampling strategy.

In order to better describe the advantages of the linear sampling strategy, we introduce the deviation score $(D-Score)$. We calculate the difference between the performance of each sampling strategy under the current result and the best result, and calculate the ratio of this difference to the best result. For a certain sampling strategy, the weighted average of the ratio calculated by all comparison models and all evaluation indicators is the deviation score of the sampling strategy. The lower the score is, the smaller the deviation between the sampling strategy and the optimal effect, that is, the stronger the ability to achieve the optimal result under all circumstances.

\begin{equation}
    D-Score = \frac{w}{n} \sum_{i=1}^{n} \frac{R_{best}-R_{current}}{R_{best}}
\end{equation}

Here $n$ refers to the number of models involved in comparison, and $w$ is the weight of an evaluation metric. For the selected network model, the best result of an evaluation metric is $R_{best}$ under different sampling strategies, and $R_{current}$ is the evaluation result under current sampling strategy.

We calculated $D-Score$ for each of the 8 sampling strategies, and the results are shown in Tab. \ref{tab:linear}. Among all the sampling strategies, the linear sampling has the lowest $D-Score$, which is the best performance, indicating that in most cases their evaluation results have little deviation from the optimal results, which means that this strategy can be regarded as excellent and stable performance, and will not lead to the degradation of the performance of the network.

\begin{figure}[ht]
\centerline{\includegraphics[width=\linewidth]{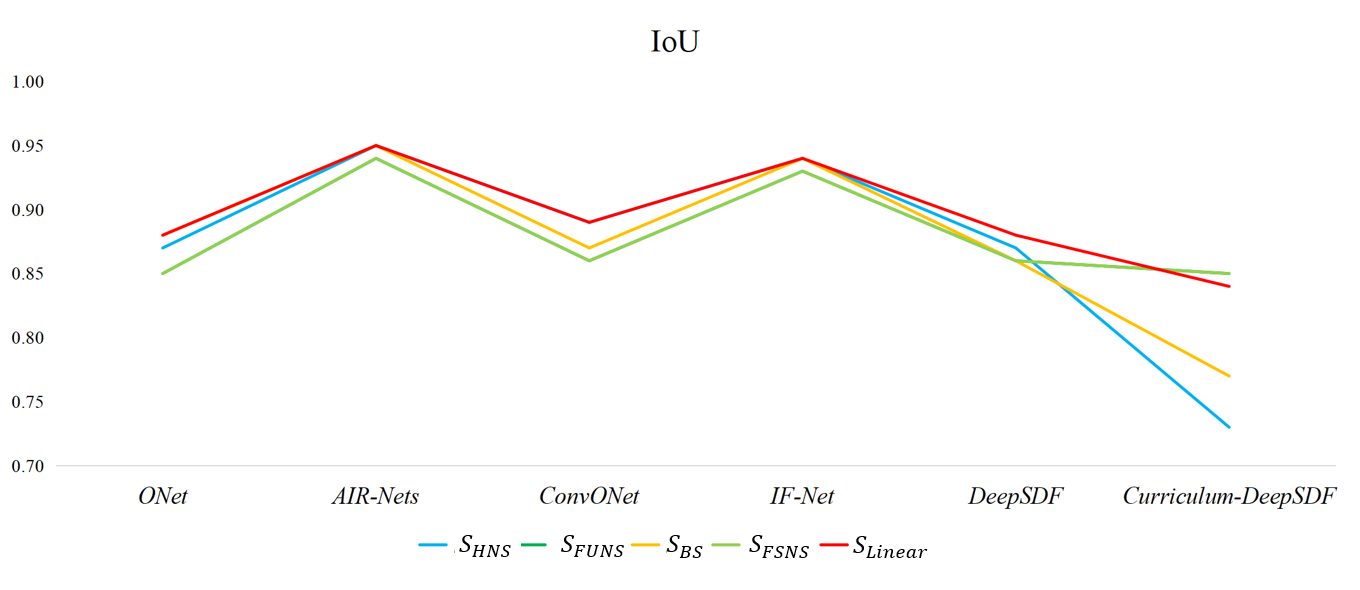}}
\caption{Evaluation results of several sampling strategies under the IoU metric.}
\label{fig:linear_IOU}
\end{figure}

We more intuitively demonstrate the advantages of $S_{Linear}$ in Fig. \ref{fig:linear_IOU}. For several strategies with low $D-Score$, the evaluation results of IoU under different models were plotted as line plots. $S_{Linear}$ (red) achieved the best results on almost all models, and there was almost no significant decline relative to other sampling strategies.

To sum up, the linear sampling strategy proposed by us has two major advantages: first, its performance is not inferior to that of other sampling strategies. It shows remarkable capability in almost all models, and the evaluation results are often the best or almost the same as the best results. Second, strong robustness, strong adaptability to different types of network models, will not appear incompatible with a certain network phenomenon.

\subsection{The Sampling Strategy based on Distance Mask}

\subsubsection{Algorithm}

\begin{figure}[h]
\centerline{\includegraphics[width=\linewidth]{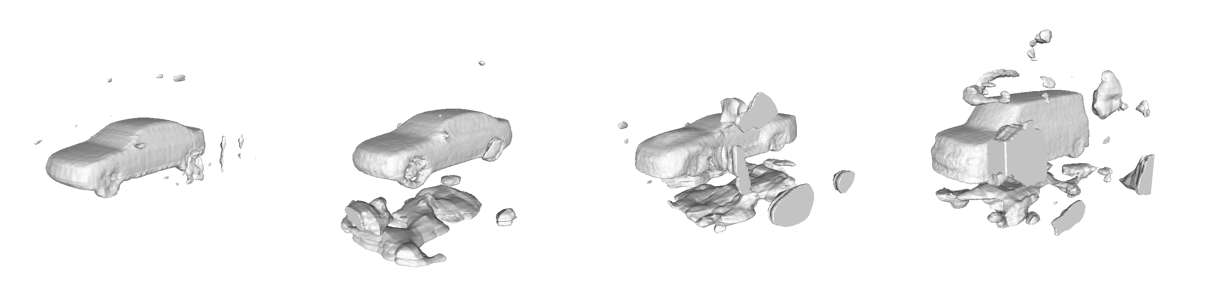}}
\caption{Some cases of implicit 3D reconstruction networks' failure.}
\label{fig:error}
\end{figure}

We find that for implicit neural representation, when the reconstruction effect of the network is poor, errors tend to appear on the far surface of the 3D object, as shown in Fig. \ref{fig:error}. In other words, when different sampling strategies are adopted, the neural network can usually successfully learn how to predict the spatial attributes of query points at the near surface position, but fail to learn at the far surface position.

Based on this phenomenon, an assumption is made that if the neural network is prone to prediction errors for query points at far surface locations, these failures can be avoided by shielding this part of query points, so that the model performance can be maintained at the optimal level.

Therefore, we propose a sampling strategy based on distance mask.  Each point in the set of query points is assigned a mask according to its distance from the surface of the 3D object, and then the neural network determines whether the point will participate in the training of the model according to the mask of the query point.  Specifically, in the boundary space, we set a distance range, query points within this range relative to the 3D object surface will be retained and transmitted to the neural network for forward propagation, while query points far away from this range will be directly abandoned and no longer participate in the calculation of the neural network, and the prediction attribute of these far surface query points will be set as predetermined values.

It should be noted that the spatial query point at the far surface location is not only outside the 3D object, but also inside the 3D object.   However, in the generation stage of implicit neural representation, the shape input provided for the neural network is unstructured and incomplete point cloud.   For a query point at a random location in space, it is difficult to distinguish between the inside and outside of the point cloud.   In other words, during the generation phase, it is difficult to provide complex predefined spatial attributes for far-surface query points, and the Unsigned Distance Function (UDF) as the spatial property of query points is suitable for our distance mask algorithm, because invalid points under this implicit function can be regarded as having only one category (e.g. UDF $>$ 0.1).

\begin{figure}[h]
\centerline{\includegraphics[width=\linewidth]{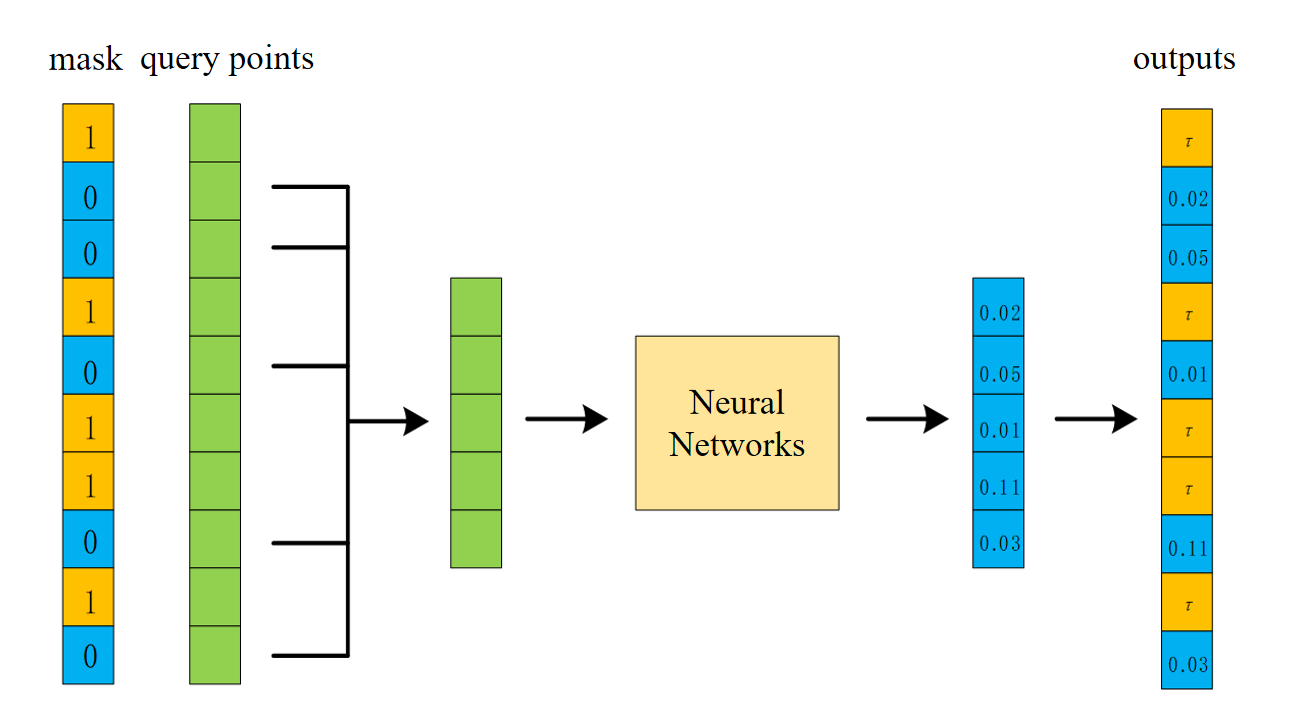}}
\caption{The implementation of the distance mask.}
\label{fig:mask}
\end{figure}

\subsubsection{Evaluation}

In this section, we adopt the distance mask algorithm with threshold $\tau = 0.1$. Since the distance mask algorithm is only applicable to UDF at present, we set the implicit function of the model as UDF in the following experiment. Considering that this implicit function cannot generate mesh shapes by extracting zero level sets, we followed the dense surface point cloud extraction algorithm proposed by Chibane \textit{et al.} \cite{ndf} in the generation phase.

Here, we verify the improvement ability of the distance mask algorithm under the near-surface sampling strategy and the uniform sampling strategy on ConvONet\cite{convonet}. Tab. \ref{tab:mask} shows the comparison of evaluation results of the model before and after using the distance mask.

\begin{table}[htb]   
    \caption{The enhancement effect of ConvONet with distance mask.}
    \begin{center}
    
    \begin{tabular}{|c|c|c|}
    \hline
    ConvONet & $Chamfer-L_2$ & $F-Score(1.5\%)$ \\
    \hline
    $S_{NS}$ & $1.2\times10^{-2}$ & 0.67\\
    $Mask-S_{NS}$ & $1.4 \times 10^{-4}$ & 0.92\\
    \hline
    $S_{UNI}$ & $9.1\times10^{-4}$ & 0.49\\
    $Mask-S_{UNI}$ & $3.4 \times 10^{-4}$ & 0.68\\
    \hline
    \end{tabular}
    \end{center}
    
\label{tab:mask}
\end{table}

Obviously, the distance mask has a very significant effect on the model. When the distance mask was applied to $S_{NS}$, the $F-Score(1.5\%)$ metric improved by $37\%$ and the $Chamfer-L_2$ metric improved by two orders of magnitude. Distance mask also has a significant improvement effect on $S_{UNI}$, with $F-Score(1.5\%)$ and $Chamfer-L_2$ increasing by $39\%$ and $63\%$ respectively.

\begin{figure}[h]
\centerline{\includegraphics[width=\linewidth]{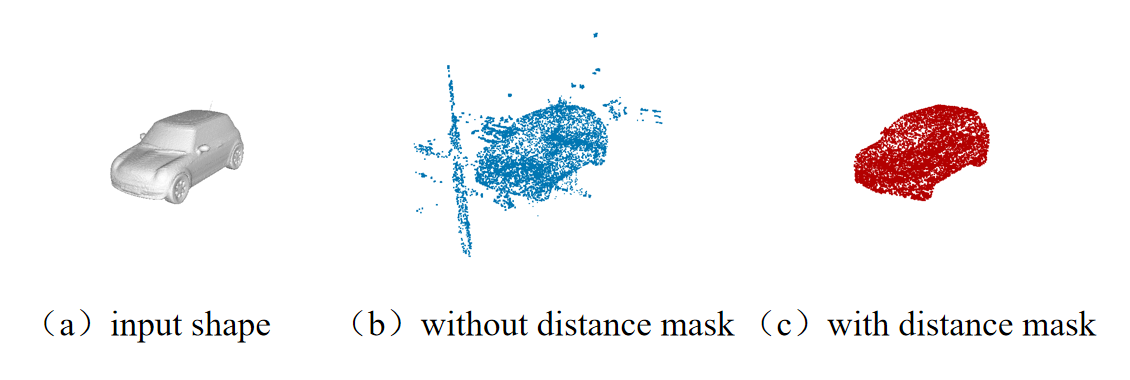}}
\caption{Improved effect of the distance mask method.}
\label{fig:mask_recons}
\end{figure}

In order to visually demonstrate the improvement of the diatance mask algorithm on ConvONet under the near surface sampling strategy, 10,000 points were sampled from the shaped surface containing 1,000,000 dense point clouds extracted by the neural network and displayed in three dimensions, as shown in Fig. \ref{fig:mask_recons}. Without diatance mask, it is difficult for the network to predict the real UDF value at the far surface position, and the extracted point cloud surface is distributed discreteously. The diatance mask operation ensures the accurate reconstruction of 3D shape to a large extent, and the point cloud surface is concentrated near the real shape contour surface.

In summary, the distance mask algorithm ensures network performance by forcing correction of UDF values of far surface position query points, which helps to avoid poor model performance due to inappropriate sampling strategies.  When the compatibility between the sampling strategy and the implicit 3D reconstruction network cannot be ensured, the distance mask algorithm is considered to be effective.

\section{Conclusion}
In this work, we discuss the sampling strategy of query points in implicit 3D reconstruction. On the one hand, we explore the deep reasons for how sampling strategies influence the performance of implicit 3D reconstruction from three aspects. The network type is the core reason, and different types of networks should have different emphases when sampling query points. The classification implicit function has a higher tolerance to changes in sampling strategies, and the importance of sampling density also cannot be ignored. On the other hand, we explore how to make more reasonable use of the effective points in the query points, and propose two sampling strategy algorithms with more universality and robustness: linear sampling and distance mask.

In the future, we will continue to push forward the insufficient part of this work in two directions. Firstly, we focus more on the overall evaluation trend among sampling strategies of implicit 3D reconstruction, and hope to continue to explore the relationship between sampling strategies and specific evaluation metric in the future. Secondly, our distance mask algorithm is only applicable to the unsigned distance function, which will be improved and further extended in the future to adapt to more implicit functions.

\section*{Acknowledgment}
This work is supported by Jilin University (Grant No.419021422B08).

\bibliographystyle{ieeetr}
\bibliography{reference}

\begin{thebibliography}{10}

\bibitem{onet}
L.~Mescheder, M.~Oechsle, M.~Niemeyer, S.~Nowozin, and A.~Geiger, ``Occupancy
  networks: Learning 3d reconstruction in function space,'' in {\em Proceedings
  of the IEEE/CVF conference on computer vision and pattern recognition},
  pp.~4460--4470, 2019.

\bibitem{convonet}
S.~Peng, M.~Niemeyer, L.~Mescheder, M.~Pollefeys, and A.~Geiger,
  ``Convolutional occupancy networks,'' in {\em European Conference on Computer
  Vision}, pp.~523--540, Springer, 2020.

\bibitem{if-net}
J.~Chibane, T.~Alldieck, and G.~Pons-Moll, ``Implicit functions in feature
  space for 3d shape reconstruction and completion,'' in {\em Proceedings of
  the IEEE/CVF Conference on Computer Vision and Pattern Recognition},
  pp.~6970--6981, 2020.

\bibitem{ndf}
J.~Chibane, G.~Pons-Moll, {\em et~al.}, ``Neural unsigned distance fields for
  implicit function learning,'' {\em Advances in Neural Information Processing
  Systems}, vol.~33, pp.~21638--21652, 2020.

\bibitem{air-nets}
S.~Giebenhain and B.~Goldl{\"u}cke, ``Air-nets: An attention-based framework
  for locally conditioned implicit representations,'' in {\em 2021
  International Conference on 3D Vision (3DV)}, pp.~1054--1064, IEEE, 2021.

\bibitem{deepsdf}
J.~J. Park, P.~Florence, J.~Straub, R.~Newcombe, and S.~Lovegrove, ``Deepsdf:
  Learning continuous signed distance functions for shape representation,'' in
  {\em Proceedings of the IEEE/CVF conference on computer vision and pattern
  recognition}, pp.~165--174, 2019.

\bibitem{c-deepsdf}
Y.~Duan, H.~Zhu, H.~Wang, L.~Yi, R.~Nevatia, and L.~J. Guibas, ``Curriculum
  deepsdf,'' in {\em European Conference on Computer Vision}, pp.~51--67,
  Springer, 2020.

\bibitem{pifu}
S.~Saito, Z.~Huang, R.~Natsume, S.~Morishima, A.~Kanazawa, and H.~Li, ``Pifu:
  Pixel-aligned implicit function for high-resolution clothed human
  digitization,'' in {\em Proceedings of the IEEE/CVF International Conference
  on Computer Vision}, pp.~2304--2314, 2019.

\bibitem{work6}
R.~Venkatesh, T.~Karmali, S.~Sharma, A.~Ghosh, R.~V. Babu, L.~A. Jeni, and
  M.~Singh, ``Deep implicit surface point prediction networks,'' in {\em
  Proceedings of the IEEE/CVF International Conference on Computer Vision
  (ICCV)}, pp.~12653--12662, October 2021.

\bibitem{voxel}
C.~B. Choy, D.~Xu, J.~Gwak, K.~Chen, and S.~Savarese, ``3d-r2n2: A unified
  approach for single and multi-view 3d object reconstruction,'' in {\em
  Proceedings of the European Conference on Computer Vision ({ECCV})}, 2016.

\bibitem{mesh}
Z.~Chen and H.~Zhang, ``Learning implicit fields for generative shape
  modeling,'' in {\em Proceedings of the IEEE/CVF Conference on Computer Vision
  and Pattern Recognition (CVPR)}, June 2019.

\bibitem{pointnet}
C.~R. Qi, H.~Su, K.~Mo, and L.~J. Guibas, ``Pointnet: Deep learning on point
  sets for 3d classification and segmentation,'' in {\em Proceedings of the
  IEEE Conference on Computer Vision and Pattern Recognition (CVPR)}, July
  2017.

\bibitem{pointnet++}
C.~R. Qi, L.~Yi, H.~Su, and L.~J. Guibas, ``Pointnet++: Deep hierarchical
  feature learning on point sets in a metric space,'' in {\em Advances in
  Neural Information Processing Systems} (I.~Guyon, U.~V. Luxburg, S.~Bengio,
  H.~Wallach, R.~Fergus, S.~Vishwanathan, and R.~Garnett, eds.), vol.~30,
  Curran Associates, Inc., 2017.

\bibitem{mcubes}
W.~E. Lorensen and H.~E. Cline, ``Marching cubes: A high resolution 3d surface
  construction algorithm,'' {\em ACM siggraph computer graphics}, vol.~21,
  no.~4, pp.~163--169, 1987.

\bibitem{im-net}
Z.~Chen and H.~Zhang, ``Learning implicit fields for generative shape
  modeling,'' in {\em Proceedings of the IEEE/CVF Conference on Computer Vision
  and Pattern Recognition}, pp.~5939--5948, 2019.

\bibitem{disn}
Q.~Xu, W.~Wang, D.~Ceylan, R.~Mech, and U.~Neumann, ``Disn: Deep implicit
  surface network for high-quality single-view 3d reconstruction,'' {\em
  Advances in Neural Information Processing Systems}, vol.~32, 2019.

\bibitem{work1}
K.~Genova, F.~Cole, A.~Sud, A.~Sarna, and T.~Funkhouser, ``Local deep implicit
  functions for 3d shape,'' in {\em Proceedings of the IEEE/CVF Conference on
  Computer Vision and Pattern Recognition (CVPR)}, June 2020.

\bibitem{work2}
C.~M. Jiang, A.~Sud, A.~Makadia, J.~Huang, M.~Niessner, and T.~Funkhouser,
  ``Local implicit grid representations for 3d scenes,'' in {\em Proceedings of
  the IEEE/CVF Conference on Computer Vision and Pattern Recognition (CVPR)},
  June 2020.

\bibitem{work3}
M.~Li and H.~Zhang, ``D2im-net: Learning detail disentangled implicit fields
  from single images,'' in {\em Proceedings of the IEEE/CVF Conference on
  Computer Vision and Pattern Recognition (CVPR)}, pp.~10246--10255, June 2021.

\bibitem{work4}
Z.~Chen, Y.~Zhang, K.~Genova, S.~Fanello, S.~Bouaziz, C.~H\"ane, R.~Du,
  C.~Keskin, T.~Funkhouser, and D.~Tang, ``Multiresolution deep implicit
  functions for 3d shape representation,'' in {\em Proceedings of the IEEE/CVF
  International Conference on Computer Vision (ICCV)}, pp.~13087--13096,
  October 2021.

\bibitem{work5}
J.~Tang, J.~Lei, D.~Xu, F.~Ma, K.~Jia, and L.~Zhang, ``Sa-convonet:
  Sign-agnostic optimization of convolutional occupancy networks,'' in {\em
  Proceedings of the IEEE/CVF International Conference on Computer Vision
  (ICCV)}, pp.~6504--6513, October 2021.

\bibitem{work7}
E.~Tretschk, A.~Tewari, V.~Golyanik, M.~Zollh{\"o}fer, C.~Stoll, and
  C.~Theobalt, ``Patchnets: Patch-based generalizable deep implicit 3d shape
  representations,'' in {\em Computer Vision -- ECCV 2020} (A.~Vedaldi,
  H.~Bischof, T.~Brox, and J.-M. Frahm, eds.), (Cham), pp.~293--309, Springer
  International Publishing, 2020.

\bibitem{work8}
T.~Takikawa, J.~Litalien, K.~Yin, K.~Kreis, C.~Loop, D.~Nowrouzezahrai,
  A.~Jacobson, M.~McGuire, and S.~Fidler, ``Neural geometric level of detail:
  Real-time rendering with implicit 3d shapes,'' in {\em Proceedings of the
  IEEE/CVF Conference on Computer Vision and Pattern Recognition (CVPR)},
  pp.~11358--11367, June 2021.

\bibitem{work9}
M.~Ibing, I.~Lim, and L.~Kobbelt, ``3d shape generation with grid-based
  implicit functions,'' in {\em Proceedings of the IEEE/CVF Conference on
  Computer Vision and Pattern Recognition (CVPR)}, pp.~13559--13568, June 2021.

\bibitem{work10}
C.~Stucker, B.~Ke, Y.~Yue, S.~Huang, I.~Armeni, and K.~Schindler,
  ``{ImpliCity}: City modeling from satellite images with deep implicit
  occupancy fields,'' {\em {ISPRS} Annals of the Photogrammetry, Remote Sensing
  and Spatial Information Sciences}, vol.~V-2-2022, pp.~193--201, 2022.

\bibitem{rgbd}
D.~Azinovi\'c, R.~Martin-Brualla, D.~B. Goldman, M.~Nie{\ss}ner, and J.~Thies,
  ``Neural rgb-d surface reconstruction,'' in {\em Proceedings of the IEEE/CVF
  Conference on Computer Vision and Pattern Recognition (CVPR)},
  pp.~6290--6301, June 2022.

\bibitem{resnet}
K.~He, X.~Zhang, S.~Ren, and J.~Sun, ``Deep residual learning for image
  recognition,'' in {\em Proceedings of the IEEE conference on computer vision
  and pattern recognition}, pp.~770--778, 2016.

\bibitem{3dldm}
G.~Nam, M.~Khlifi, A.~Rodriguez, A.~Tono, L.~Zhou, and P.~Guerrero, ``3d-ldm:
  Neural implicit 3d shape generation with latent diffusion models,'' {\em
  arXiv preprint arXiv:2212.00842}, 2022.

\bibitem{gifs}
J.~Ye, Y.~Chen, N.~Wang, and X.~Wang, ``Gifs: Neural implicit function for
  general shape representation,'' in {\em Proceedings of the IEEE/CVF
  Conference on Computer Vision and Pattern Recognition}, pp.~12829--12839,
  2022.

\bibitem{geoudf}
S.~Ren, J.~Hou, X.~Chen, Y.~He, and W.~Wang, ``Geoudf: Surface reconstruction
  from 3d point clouds via geometry-guided distance representation,'' 2022.

\bibitem{neuralmeshing}
M.~Vetsch, S.~Lombardi, M.~Pollefeys, and M.~R. Oswald, ``Neuralmeshing:
  Differentiable meshing of implicit neural representations,'' in {\em Pattern
  Recognition: 44th DAGM German Conference, DAGM GCPR 2022, Konstanz, Germany,
  September 27--30, 2022, Proceedings}, pp.~317--333, Springer, 2022.

\bibitem{importance}
T.~Davies, D.~Nowrouzezahrai, and A.~Jacobson, ``Overfit neural networks as a
  compact shape representation,'' 2020.

\bibitem{mesh_fusion}
D.~Stutz and A.~Geiger, ``Learning 3d shape completion under weak
  supervision,'' {\em CoRR}, vol.~abs/1805.07290, 2018.

\end{thebibliography}

\end{document}